\documentclass[10pt,twocolumn,letterpaper]{article}

\usepackage{iccv}
\usepackage{times}
\usepackage{epsfig}
\usepackage{graphicx}
\usepackage{amsmath}
\usepackage{amssymb}

\usepackage{stmaryrd}
\usepackage{dsfont}
\usepackage{textcomp}

\usepackage[pagebackref=true,breaklinks=true,letterpaper=true,colorlinks,bookmarks=false]{hyperref}

\iccvfinalcopy 

\def\iccvPaperID{710} 

\ificcvfinal\fi
\begin{document}

\title{Open Vocabulary Scene Parsing}

\author{Hang Zhao$^{1}$, Xavier Puig$^{1}$, Bolei Zhou$^{1}$, Sanja Fidler$^{2}$, Antonio Torralba$^{1}$\\
$^{1}$Massachusetts Institute of Technology, USA\\
$^{2}$University of Toronto, Canada\\
}

\maketitle

\begin{abstract}
Recognizing arbitrary objects in the wild has been a challenging problem due to the limitations of existing classification models and datasets.
In this paper, we propose a new task that aims at parsing scenes with a large and open vocabulary, and several evaluation metrics are explored for this problem. Our proposed approach to this problem is a joint image pixel and word concept embeddings framework, where word concepts are connected by semantic relations. 
We validate the open vocabulary prediction ability of our framework on ADE20K dataset which covers a wide variety of scenes and objects. We further explore the trained joint embedding space to show its interpretability.
\end{abstract}

\section{Introduction}

One of the grand goals in computer vision is to recognize and segment arbitrary objects in the wild. Recent efforts in image classification/detection/segmentation have shown this trend: emerging image datasets enable recognition on a large-scale ~\cite{deng2009imagenet,xiao2010sun,zhou2014learning}, while image captioning can be seen as a special instance of this task~\cite{karpathy2015deep}. However, nowadays most recognition models are still not capable of classifying objects at the level of a human, in particular, taking into account the taxonomy of object categories. Ordinary people or laymen classify things on the entry-levels, and experts give more specific labels: there is no object with a single correct label, so the prediction vocabulary is inherently open-ended.
Furthermore, there is no widely-accepted way to evaluate open-ended recognition tasks, which is also a main reason this direction is not pursued more often.

\begin{figure}
\centering
\includegraphics[width=0.4\textwidth]{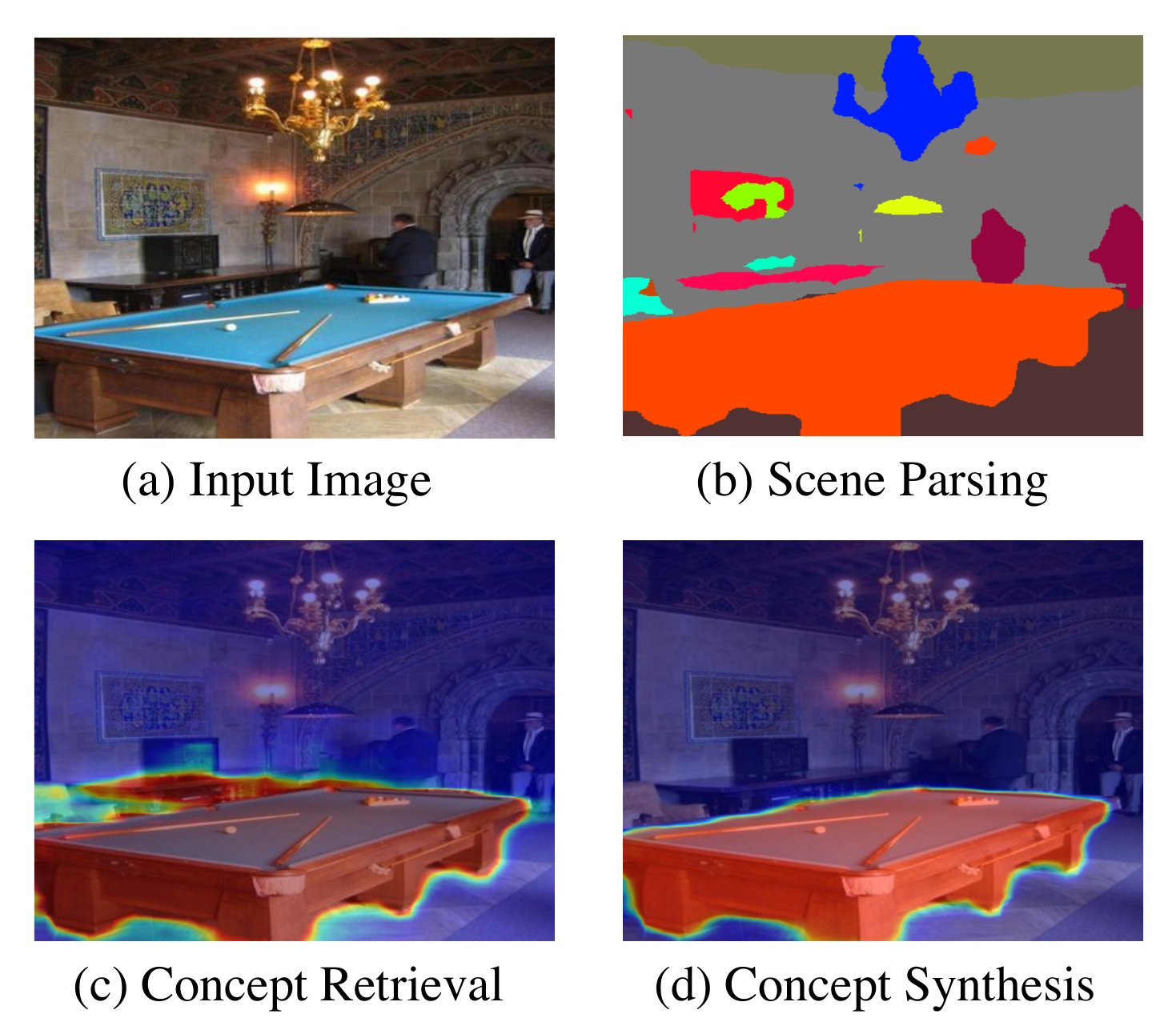}
\caption{We propose an open vocabulary framework such that given (a) an input image, we can perform (b) scene parsing, (c) concept retrieval ("table"), and (d) concept synthesis (both ``game equipment" and ``table") through arithmetic operations in the joint image-concept embedding space.}
\label{fig:teaser}
\vspace{-5px}
\end{figure}

In this work, we are pushing towards open vocabulary scene parsing: model predictions are not limited to a fixed set of categories, but also words in a larger dictionary, or even a knowledge graph. Considering existing image parsing datasets only contain a small number of categories  (\texttildelow 100 classes), there is much more a model can learn from those images given extra semantic knowledge, like WordNet dictionary (\texttildelow 100,000 synsets) or \textit{Word2Vec} from external corpus.

To solve this new problem, we propose a framework that is able to segment all objects in an image 
using open vocabulary labels. In particular, while the method strives to label each pixel with the same word as the one used by the human annotator, it resorts to a taxonomy when it is not sure about its prediction.
As a result, our model aims to make plausible predictions even for categories that have not been shown during training, \textit{e.g.} if the model has never seen \textit{tricycle}, it may still give a confident guess on \textit{vehicle}, performing more like a human.

Our framework incorporates hypernym/hyponym relations from WordNet \cite{miller1995wordnet} to help parsing. More concretely, word concepts and image pixel features are embedded into a joint high-dimentional vector space so that (1) hypernym/hyponym relations are preserved for the concepts, (2) image pixel embeddings are close to concepts related to their annotations according to some distance measures. This framework offers three major advantages: (1) predictions are made in a structured way, \textit{i.e.}, they can be intermediate nodes in WordNet, and thus yielding more reasonable mistakes; (2) it is an end-to-end trainable system, its vocabulary can be huge and is easily extensible; (3) the framework leaves more freedom to the annotations: inconsistent annotations from workers with different domain knowledge have less of an affect on the performance of the model. 

We additionally explored several evaluation metrics, which are useful measures not only for our open vocabulary parsing tasks, but also for any large-scale recognition tasks where confusions often exist.

The open vocabulary parsing ability of the proposed framework is evaluated on the recent ADE20K dataset \cite{zhou2016semantic}. We further explore the properties of the embedding space by concept retrieval, classification boundary loosing and concept synthesis with arithmetics.

\subsection{Related work}
Our work is related to different topics in literature which
we briefly review below.

\textbf{Semantic segmentation and scene parsing}. Due to astonishing performance of deep learning, in particular CNNs~\cite{krizhevsky2012imagenet}, 
pixel-wise dense labeling has received significant amount of attention. Existing work include fully convolutional neural network (FCN) \cite{long2015fully}, deconvolutional neural network \cite{noh2015learning}, encoder-decoder SegNet \cite{badrinarayanan2015segnet}, dilated neural network \cite{CP2016Deeplab, YuKoltun2016}, \textit{etc}.

These networks perform well on datasets like PASCAL VOC~\cite{everingham2010pascal} with 20 object categories, Cityscapes~\cite{Cordts2015Cvprw} with 30 classes, and a recently released benchmark SceneParse150~\cite{zhou2016semantic} covering 150 most frequent daily objects. However, these models are not easily adaptable to new objects. In this paper we aim at going beyond this limit and to make predictions in the wild. 

\textbf{Zero-shot learning}. Zero-shot learning addresses knowledge transfer and generalization \cite{rohrbach2011evaluating,guillaumin2012large}. Models are often evaluated on unseen categories, and predictions are made based on the knowledge extracted from the training categories. Rohrbach \cite{rohrbach2010helps} introduced the idea to transfer large-scale linguistic knowledge into vision tasks. Socher \textit{et~al.} \cite{socher2013zero} and Frome \textit{et~al.} \cite{frome2013devise} directly embedded visual features into the word vector space so that visual similarities are connected to semantic similarities. Norouzi \textit{et~al.} \cite{norouzi2013zero} used a convex combination of visual features of training classes to represent new categories. Attribute-based methods are another major direction in zero-shot learning that maps object attribute labels or language descriptions to visual classifiers~\cite{parikh2011relative,Akata_2013_CVPR,lei2015predicting,lampert2014attribute}.

\textbf{Hierarchical classifications}. Hierarchical classification addresses the common circumstances that candidate categories share hierarchical semantic relations. Zweig \textit{et~al.} \cite{zweig2007exploiting} combined classifiers on different levels to help improve classification. Deng \textit{et~al.}~\cite{deng2012hedging} achieved hierarchical image-level classification by trading off accuracy and gain as an optimization problem. Ordonez \textit{et~al.}~\cite{ordonez2013large}, on the other hand, proposed to make entry-level predictions when dealing with a large number of categories. More recently, Deng \textit{et~al.}~\cite{deng2014large} formulated a label relation graph that could be directly integrated with deep neural networks.

Our approach on hierarchical parsing is inspired by the order-embeddings work~\cite{vendrov2015order}, we attempt to construct an asymmetric embedding space, so that both image features and hierarchical information in the knowledge graph are effectively and implicitly encoded by the deep neural networks.
While most previous approaches combine deep neural networks with optimization methods like conditional random fields so that the semantic relatedness is incorporated into the framework, the advantage of our approach is that it makes an end-to-end trainable network, which is easily scalable when dealing with larger datasets in practical applications.


\begin{figure}
\centering
\includegraphics[width=0.43\textwidth]{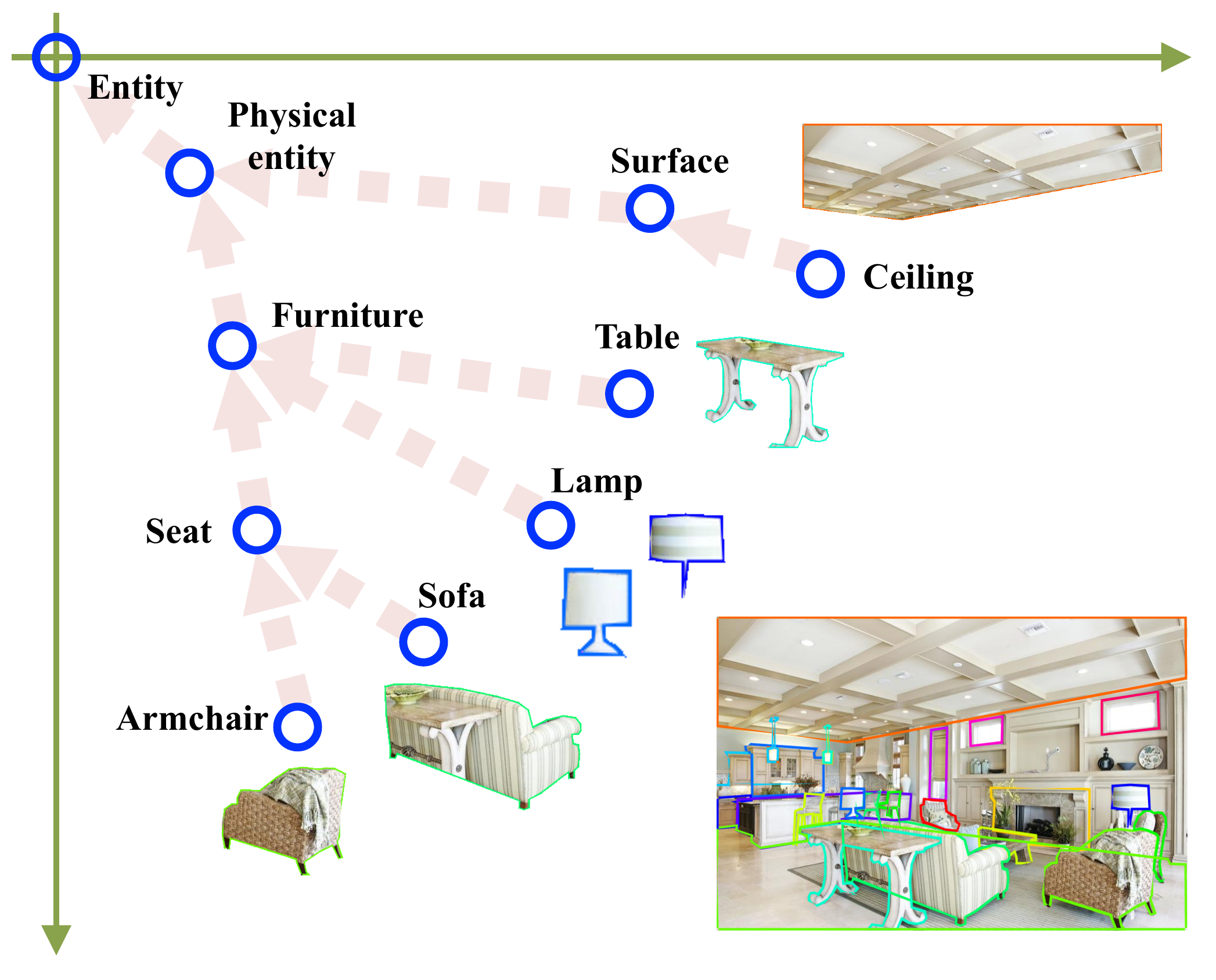}
\caption{Jointly embedding vocabulary concepts and image pixel features.}
\label{fig:order}
\end{figure}
\begin{figure*}
\centering
\includegraphics[width=0.9\textwidth]{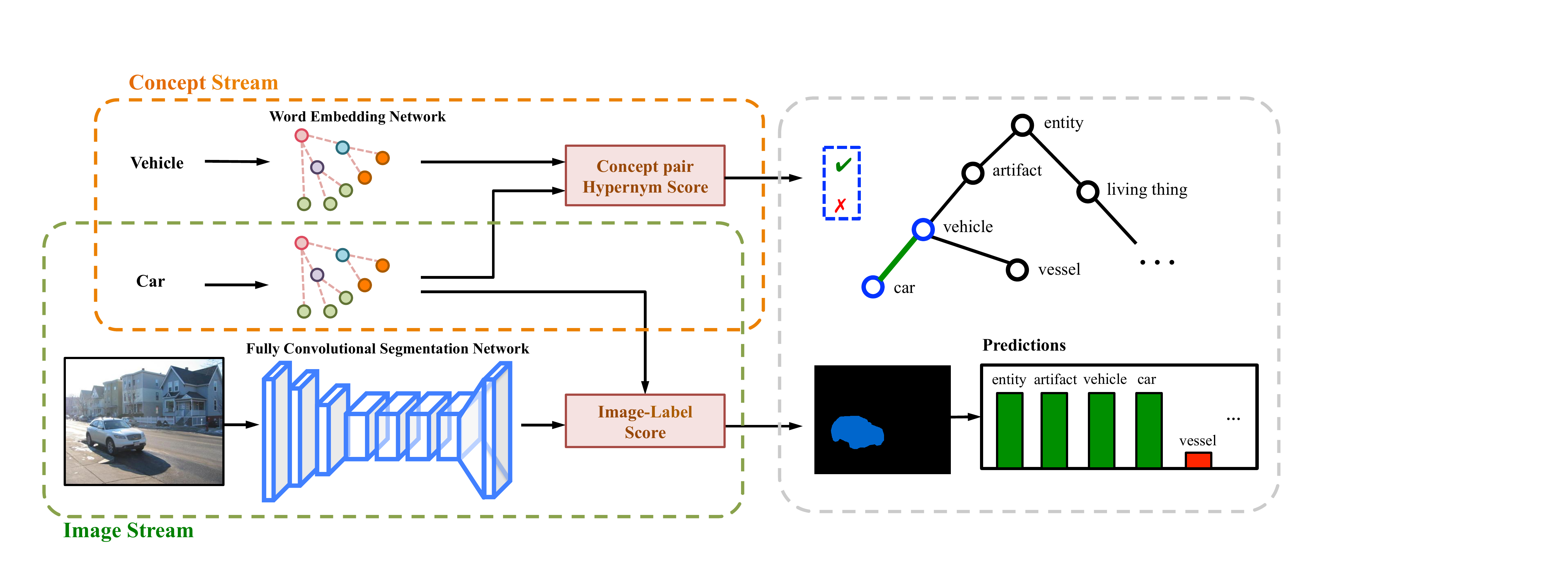}
\caption{The open vocabulary parsing network. The concept stream encodes the word concept hierarchy based on dictionaries like WordNet. The image stream parses the images based on the learned hierarchy.}
\label{fig:network}
\end{figure*}
\section{Learning joint embeddings for pixel features and word concepts}
We treat open-ended scene parsing as a retrieval problem for each pixel, following the ideas of image-caption retrieval work~\cite{vendrov2015order}.
Our goal is to embed image pixel features and word concepts  into a joint high-dimensional positive vector space $\mathds{R}_+^N$, as illustrated in Figure ~\ref{fig:order}. The guiding principle while constructing the joint embedding space is that image features should be close to their concept labels, and word concepts should preserve their semantic hypernym/hyponym relations. In this embedding space, (1) vectors close to origin are general concepts, and vectors with larger norms represent higher specificity; (2) hypernym/hyponym relation is defined by whether one vector is smaller/greater than another vector in all the $N$ dimensions. A hypernym scoring function is crucial in building this embedding space, which will be detailed in Section \ref{sec:scoring}.

Figure ~\ref{fig:network} gives an overview of our proposed framework. It is composed of two streams: a concept stream and an image stream. The concept stream tries to encode the pre-defined semantics: it learns an embedding function $f(\cdot)$ that maps the words into $\mathds{R}_+^N$ so that they preserve the hypernym/hyponym relationship between word concepts. The image stream $g(\cdot)$ embeds image pixels into the same space by pushing them close to their labels (word concepts). We describe these two streams in more details in Section \ref{sec:concept_stream} and \ref{sec:image_stream}.

\subsection{Scoring functions}
\label{sec:scoring}
In this embedding problem, training is performed on pairs: image-label pairs and concept-concept pairs. For either of the streams, the goal is to maximize scores of matching pairs and minimize scores of non-matching pairs. So the choice of scoring functions $S(x,y)$ becomes important. There are symmetric scoring functions like $L_p$ distance and cosine similarity widely used in the embedding tasks,
\begin{equation}
	S_{Lp}(x,y) = -\|x-y\|^p, \quad
    S_{cos}(x,y) = x \cdot y.
\label{eq:scoring-symm}
\end{equation}
In order to reveal the asymmetric hypernym/hyponym relations between word concepts, hypernym scoring function~\cite{vendrov2015order} is indispensable,
\begin{equation}
	S_{hyper}(x,y) = -\| max(0, x-y)\|^p.
\label{eq:scoring-hyper}
\end{equation}
If $x$ is hypernym of $y$ ($x \succeq y$), then ideally all the coordinates of $x$ are smaller than $y$ ($\bigwedge_i (x_i\le y_i)$), so $S_{hyper}(x,y) = S_{hyper,max} = 0$. Note that due to asymmetry, swapping $x$ and $y$ will result in total different scores.

\subsection{Concept stream}
\label{sec:concept_stream}
The objective of the concept stream is to build up semantic relations in the embedding space. In our case, the semantic structure is obtained from WordNet hypernym/hyponym relations. Consider all the vocabulary concepts form a directed acyclic graph (DAG) $H=(V,E)$, sharing a common root $\hat{v} \in V$ ``entity", each node in the graph $v \in V$ can be an abstract concept as the unions of its children nodes, or a specific class as a leaf. 
A visualization of part of the DAG we built based on WordNet and ADE20K labels can be found in Supplementary Materials. 

Internally, the concept stream include parallel layers of a shared trainable lookup table, mapping the word concepts $u,v$ to $f(u)$, $f(v)$. And then they are evaluated on hypernym scores $S_{concept}(f(u), f(v)) = S_{hyper}(f(u), f(v))$, which tells how confident $u$ is a hypernym of $v$.
A max-margin loss is used to learn the embedding function $f(\cdot)$,
\begin{gather*}
\begin{aligned}
\mathcal{L}_{concept}&(u,v)=\\
&
\begin{cases}
-S_{concept}(f(u), f(v)) & \text{if } u \succeq v,\\
\text{max}\{0,\alpha+S_{concept}(f(u), f(v))\}  & \text{otherwise}\\
\end{cases}
\end{aligned}
\end{gather*}
Note that positive samples $u \succeq v$ are the cases where $u$ is an ancestor of $v$ in the graph, so all the coordinates of $f(v)$ are pushed towards values larger than $f(u)$; negative samples can be inverted pairs or random pairs, the loss function pushes them apart in the embedding space. In our training, we fix the root of DAG ``entity" as anchor at origin, so the embedding space stays in $\mathds{R}_+^N$. 

\subsection{Image stream}
\label{sec:image_stream}
The image stream is composed of a fully convolutional network which is commonly used in image segmentation tasks, and a lookup layer shared with the word concept stream. Consider an image pixel at position $(i,j)$ with label $x_{i,j}$, its feature $y_{i,j}$ is the top layer output of the convolutional network. Our mapping function $g(y_{i,j})$ embeds the pixel features into the same space as their label $f(x_{i,j})$, and then evaluate them with a scoring function $S_{image}(f(x_{i,j}),g(y_{i,j}))$.

As label retrieval is inherently a ranking problem, negative labels $x'_{i,j}$ are introduced in training. A max-margin ranking loss is commonly used \cite{frome2013devise} to encourage the scores of true labels be larger than negative labels by a margin, 
\begin{equation}
\begin{aligned}
	& \mathcal{L}_{image}(y_{i,j})= \\
	&\resizebox{1.0\hsize}{!}{$\sum_{x'_{i,j}} \text{max}\{0,\beta - S_{image}(f(x_{i,j}),g(y_{i,j})) + S_{image}(f(x'_{i,j}),g(y_{i,j}))$\}}.
\end{aligned}
\label{eq:loss_image}
\end{equation}
In the experiment, we use a softmax loss for all our models and empirically find better performance,

\begin{equation}
\begin{aligned}
	&\mathcal{L}_{image}(y_{i,j}) = \\
    &-\log\frac{e^{S_{image}(f(x_{i,j}),g(y_{i,j}))}}{e^{S_{image}(f(x_{i,j}),g(y_{i,j}))} + \sum_{x'_{i,j}} e^{S_{image}(f(x'_{i,j}),g(y_{i,j}))}}.
\end{aligned}
\label{eq:loss_image}
\end{equation}
This loss function is a variation of triplet ranking loss proposed in \cite{hoffer2015deep}. 

The choice of scoring function here is flexible, we can either (1) simply make image pixel features ``close" to the embedding of their labels by using symmetric scores $S_{L_p}(f(x_{i,j}), g(y_{i,j}))$, $S_{cos}(f(x_{i,j}), g(y_{i,j}))$; (2) or use asymmetric hypernym score $S_{hyper}(f(x_{i,j}), g(y_{i,j}))$. In the latter case, we treat images as specific instances or specializations of their label concepts, and labels as general abstraction of the images. 

\subsection{Joint model}

Our joint model combines the two streams via a joint loss function to preserve concept hierarchy as well as visual feature similarities. In particular, we simply weighted sum the losses of two streams $\mathcal{L} = \mathcal{L}_{image} + \lambda\mathcal{L}_{concept} (\lambda=5)$ during training. We set the embedding space dimension to $N=300$, which is commonly used in word embeddings. Training and model details are described in Section \ref{sec:implementation}.

\section{Evaluation Criteria}

To better evaluate our models, metrics for different parsing tasks are explored in this section.
\subsection{Baseline flat metrics}
While working on a limited number of classes, 
four traditional criteria are good measures of the scene parsing model performance: (1) pixel-wise accuracy: the proportion of correctly classified pixels; (2) mean accuracy: the proportion of correctly classified pixels averaged over all the classes; (3) mean IoU: the intersection-over-union between the predictions and ground-truth, averaged over all the classes; (4) weighted IoU: the IoU weighted by pixel ratio of each class.

\subsection{Open vocabulary metrics}
Given the nature of open vocabulary recognition, selecting a good evaluation criteria is non-trivial. Firstly, it should leverage the graph structure of the concepts to tell the distance of the predicted class from the ground truth.
Secondly, the evaluation should correctly represent the highly unbalanced distribution of the dataset classes, which are also common in the objects seen in nature. 

To do so, for each sample/pixel, a score $s(l,p)$ is used to measure the similarity between the ground truth label $s$ and the prediction $p$. The total accuracy is the mean score over all the samples/pixels.


\subsubsection{Hierarchical precision, recall and F-score}

Hierarchical precision, recall and F-score were also known as Wu-Palmer similarity, which was originally used for lexical selection \cite{wu1994verbs}.

For two given concepts $l$ and $p$, we define the lowest common ancestor \texttt{LCA} as the most specific concept (i.e. furthest from the root Entity) that is an hypernym of both. Then 
hierarchical precision and recall are defined by the number of common hypernyms that prediction and label have over the vocabulary hierarchy $H$, formally:
\begin{equation}
	s_{HP}(l,p) = \frac{d_\mathtt{LCA}}{d_p},
    \quad s_{HR}(l,p) = \frac{d_\mathtt{LCA}}{d_l}.
\label{eq:hp}
\end{equation}
where depth of the lowest common ancestor node $d_\mathtt{LCA}$ is the number of hypernyms in common.

Combining hierarchical precision and hierarchical recall, we get hierarchical F-score $s_{HF}(l,p)$， which defined as the depth of \texttt{LCA} node over the sum of depth of label and prediction nodes:
\begin{equation}
	s_{HF}(l,p) = \frac{2 s_{HP}(l,p) \cdot s_{HR}(l,p)}{s_{HP}(l,p) + s_{HR}(l,p)} = \frac{2 \cdot d_\mathtt{LCA}}{d_l + d_p}.
\label{eq:hf}
\end{equation}

One prominent advantage of these hierarchical metrics is they penalize predictions when they go too specific. For example, ``guitar" ($d_l$=10) and ``piano" ($d_p$=10) are all ``musical instrument" ($d_\mathtt{LCA}$=8). When ``guitar" is predicted as ``piano", $s_{HF}=\frac{2 \cdot 8}{10 + 10}=0.8$; when ``guitar" is predicted as ``musical instrument", $s_{HF}=\frac{2 \cdot 8}{10 + 8}=0.89$. It agrees with human judgment that the prediction ``musical instrument" is more \textit{accurate} than ``piano".

\begin{figure*}
\centering
\includegraphics[width=1.0\textwidth]{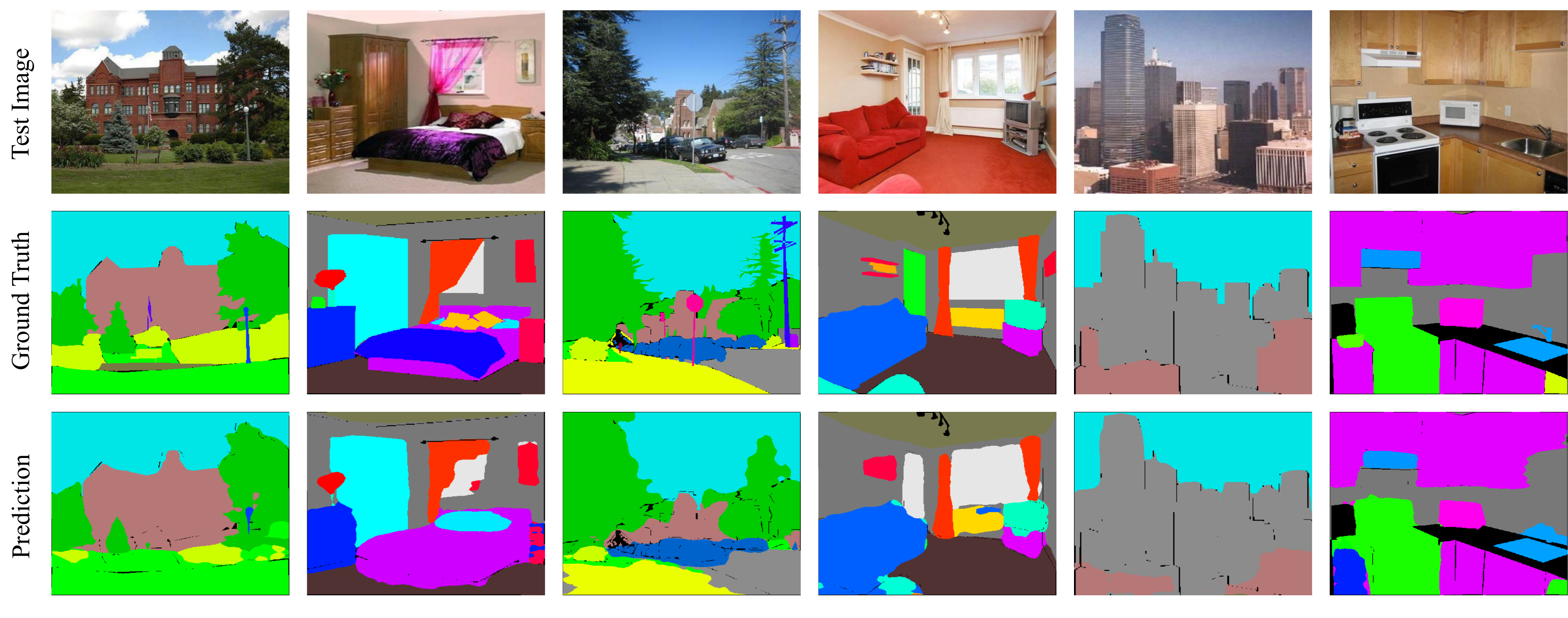}
\caption{Scene parsing results on the most frequent 150 classes, images are nearly fully segmented.}
\label{fig:result_150}
\vspace{-5px}
\end{figure*}


\vspace{-2px}
\subsubsection{Information content ratio}
\vspace{-5px}
As mentioned before, unbalanced distribution of data points could make performance dominated by frequent classes.
\textit{Information content ratio}, which was also used in lexical search, addresses these problems effectively.

According to information theory and statistics, the information content of a message is the inverse logarithm of its frequency $I(c) = -\log{P(c)}$.
We inherit this idea and pre-process our image data to get the pixel frequency of each concept $v\in H$. Specifically, the frequency of a concept is the sum of its own frequency and all its descendents' frequencies in the image dataset. 
It is expected that the root ``entity" has frequency $1.0$ and information content $0$.

During evaluations, we measure, for each testing sample, how much information our prediction gets out of the total amount of information in the label. So the final score is determined by the information of the \texttt{LCA} and that of the ground truth and predicted concepts.
\begin{equation}
 s_I(l,p) = \frac{2 \cdot I_\mathtt{LCA}}{I_{l}+I_{p}} = \frac{2 \cdot \log{P(\mathtt{LCA})}}{\log{P(l)} + \log{P(p)}}
 \label{eq:info_content_ratio}
\end{equation}
As information content ratio requires the statistics of the image dataset and the semantic hierarchy, it rewards both inference difficulty and hierarchical accuracy.

\begin{table*}
\caption{Scene parsing performance on the 150 training classes, evaluated with the baseline flat metrics.}
\label{table:eval_baseline}
\centering
\footnotesize
\begin{tabular}{| l | c | c | c | c |}
  \hline                       
  Networks &Pixel Accuracy & Mean Accuracy & Mean IoU  & Frequency Weighted IoU\\
  \hline  
Softmax \cite{zhou2016semantic}  & \textbf{73.55\%} & \textbf{44.59\%} & \textbf{0.3231} & \textbf{0.6014} \\
Conditional Softmax \cite{redmon2016yolo9000} & 72.23\% & 42.64\% & 0.3127 &  0.5942\\
Word2Vec \cite{frome2013devise} & 71.31\% & 40.31\% & 0.2918 & 0.5879 \\
\hline
Word2Vec+ & 73.11\% & 42.31\% & \textbf{0.3160} & 0.5998 \\
Image-L2 & 70.18\% & 38.89\% & 0.2174 & 0.4764 \\
Image-Cosine & 71.40\% & 40.17\% & 0.2803 & 0.5677 \\
Image-Hyper &  67.75\% & 37.10\% & 0.2158 & 0.4692 \\
Joint-L2 & 71.48\% & 39.88\% & 0.2692 & 0.5642 \\
Joint-Cosine & \textbf{73.15\%} & \textbf{43.01\%} & 0.3152 & \textbf{0.6001} \\
Joint-Hyper &  72.74\% & 42.29\% & 0.3120 & 0.5940 \\
	\hline
\end{tabular}
\vspace{-10px}
\end{table*}
\section{Experiments}
\vspace{-5px}
\subsection{Image label and concept association}
\vspace{-5px}
To learn the joint embedding, 
we associate each class in ADE20K dataset with a Synset in WordNet, representing a unique concept. The data association process requires semantic understanding, so we resort to Amazon Mechanical Turk (AMT). We develop a rigorous annotation protocol, which is detailed in Supplementary Materials.



After association, we end up with 3019 classes in the dataset having synset matches. Out of these there are 2019 unique synsets forming a DAG. All the matched synsets have \textit{entity.n.01} as the top hypernym  and there are in average 8.2 synsets in between. The depths of the ADE20K dataset annotations range from 4 to 19.

\subsection{Network implementations}
\label{sec:implementation}

\subsubsection{Concept stream}
\vspace{-5px}
The data layer of concept stream feeds the network with positive and negative vocabulary concept pairs. The positive training pairs are found by traversing the graph $H$ and find all the transitive closure hypernym pairs, \textit{e.g.} ``neckwear" and ``tie", ``clothing" and ``tie", ``entity" and ``tie"; negative samples are randomly generated before each training epoch by excluding these positive samples. Using transitive closure pair greatly improves the performance of embedding by providing us with more training data.

\vspace{-10px}
\subsubsection{Image stream}
Our core CNN in the image stream is adapted from VGG-16 by taking away \textit{pool4} and \textit{pool5} and then making all the following convolution layers dilated (or Atrous) \cite{CP2016Deeplab, YuKoltun2016}.
Considering the features of an image pixel from the last layer of the fully convolutional network \textit{fc7} to be $y_{i,j}$ with dimension $4096$, we add a $1\times 1$ convolution layer $g(\cdot)$ with weight dimension of $4096\times 300$ to embed the pixel feature. To ensure positivity, we further add a \textit{ReLU} layer.


A technique we use to improve the training is to fix the norms of the embeddings of image pixels
to be $30$, where a wide range of values will work. This technique stabilizes training numerically and speeds up the convergence. Intuitively, fixing image to have a large norm makes sense in the hierarchical embedding space: image pixels are most specific descriptions of concepts, while words are more general, and closer to the origin.

\vspace{-8px}
\subsubsection{Training and inference}
\vspace{-5px}
In all the experiments, we first train the concept stream to get the word embeddings, and then use them as initializations in the joint training.
Pre-trained weights from VGG-ImageNet \cite{Simonyan14c} are used as initializations for the image stream. 

Adam optimizer \cite{kingma2014adam} with learning rate 1e-3 is used to update weights across the model. The margin of loss functions is default to $\alpha=1.0$.

In the inference stage, there are two cases: (1) While testing on the 150 training classes, the pixel embeddings are compared with the embeddings of all the 150 candidate labels based on the scoring function, the class with the highest score is taken as the prediction; (2) While doing zero-shot predictions, on the other hand, we use a threshold on the scores to decide the cutoff score, concepts with scores above the cutoff are taken as predictions. This best threshold is found before testing on a set of 100 validation images.

\begin{figure*}
\centering
\includegraphics[width=1.0\textwidth]{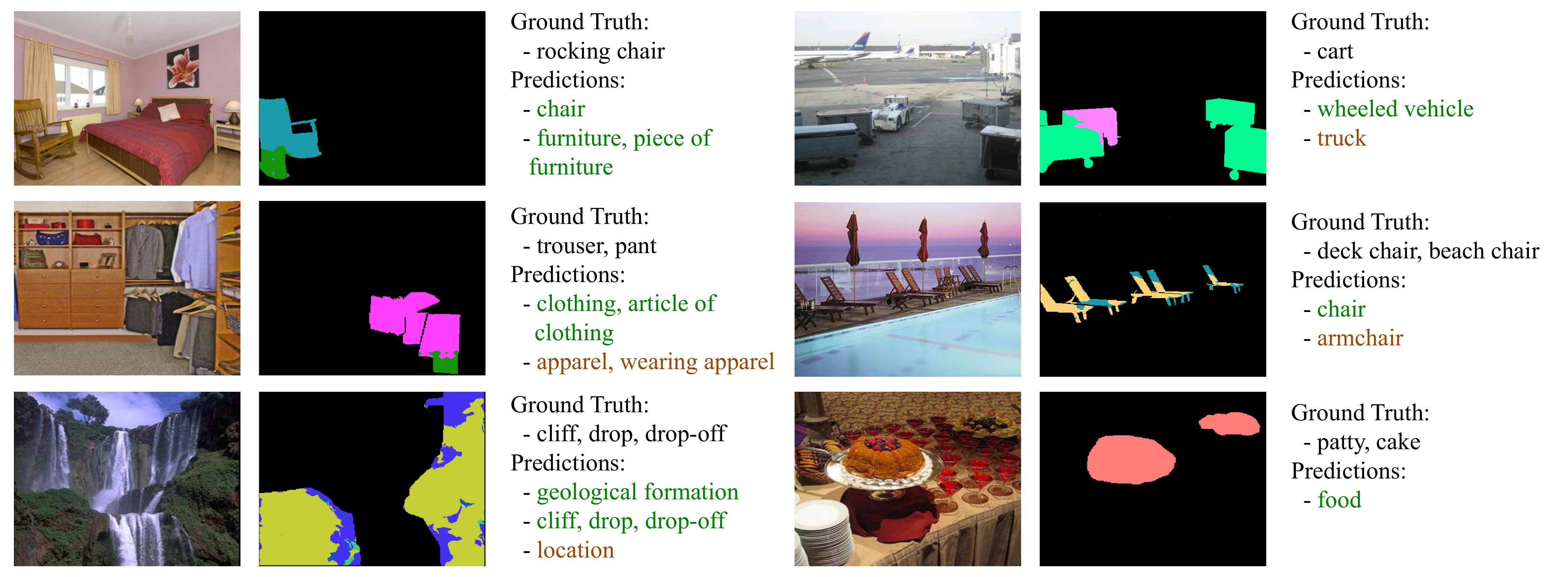}
\caption{Scene parsing results on the infrequent zero-shot object classes.}
\label{fig:result_zeroshot}
\end{figure*}

\begin{table*}
\caption{Zero-shot parsing performance, evaluated with hierarchical metrics.}
\label{table:eval_zeroshot}
\centering
\footnotesize
\begin{tabular}{| l | c | c | c | c |}
  \hline                       
  Networks & Hierarchical Precision & Hierarchical Recall & Hierarchical F-score & Information content ratio \\
  \hline  
Softmax \cite{zhou2016semantic} & 0.5620 & 0.5168 & 0.5325 & 0.1632\\
Conditional Softmax \cite{redmon2016yolo9000} & 0.5701 & 0.5146 & 0.5340 & 0.1657\\
Word2Vec \cite{frome2013devise} & 0.5782 & 0.5265 & 0.5507 & 0.1794\\
Convex Combination \cite{norouzi2013zero} & 0.5777 & 0.5384 & 0.5492 & 0.1745\\
\hline
Word2Vec+ & 0.6138 & 0.5248 & 0.5671 & 0.2002\\
Image-L2 & 0.5741 & 0.5032 & 0.5375 & 0.1650\\
Image-Hyper & 0.6318 & 0.5346 & 0.5937 & 0.2136\\
Joint-L2 & 0.5956 & 0.5385 & 0.5655 & 0.1945\\
Joint-Hyper & \textbf{0.6567} & \textbf{0.5838} & \textbf{0.6174} & \textbf{0.2226}\\
	\hline
\end{tabular}
\vspace{-5px}
\end{table*}
\subsection{Results on SceneParse150}
\label{sec:exp1}
In this section, we report the performance of our model on scene parsing task. The training is performed on the most frequent 150 classes of stuffs and objects in the ADE20K dataset, SceneParse150, where each of the class has at least $0.02\%$ of total pixels in the dataset.

We have trained some models in the references and several variants of our proposed model, all of which share the same architecture of convolutional networks to make fair comparisons. \textit{Softmax} is the baseline model that does classical multi-class classification. 

\textit{Conditional Softmax} is a hierarchical classification model proposed in \cite{redmon2016yolo9000}. It builds a tree based on the label relations, and softmax is performed only between nodes of a common parent, so only conditional probabilities for each node are computed. To get absolute probabilities during testing, the conditional probabilities are multiplied following the paths to root.

\textit{Word2Vec} is a model that simply regresses the image pixel features to pre-trained word embeddings, where we use the GoogleNews vectors. Since the dimensionality of GoogleNews vectors is 300, the weight dimension of the last convolution layer is a $1\times1\times4096\times300$. Cosine similarity and max-margin ranking loss with negative samples are used during training. This model is a direct counterpart of DeViSe\cite{frome2013devise} in our scene parsing settings.

\textit{Word2Vec+} is our improved version of \textit{Word2Vec} model. There are two major modifications: (1) We replace the max margin loss with a softmax loss as mentioned in Section \ref{sec:image_stream}; (2) We augment the GoogleNews vectors by finetuning them on domain specific corpus. Concretely, from AMT we collect 3 to 5 scene descriptive sentences for each image in the ADE20K training set (20,210 images). Then we finetune the pre-trained word vectors with skip-gram model \cite{mikolov2013distributed} for 5 epochs, and these word vectors are finally fixed for regression like \textit{Word2Vec}.

There are 6 variants of our proposed model. Model names with \textit{Image-*} refer to the cases where only image stream is trained, by fixing the concept embeddings. In models \textit{Joint-*} we train two streams together to learn a joint embedding space. Three aforementioned scoring functions are used for the image stream, their corresponding models are marked as \textit{*-L2}, \textit{*-Cosine} and \textit{*-Hyper}. 

\vspace{-10px}
\subsubsection{On the training classes}
\vspace{-5px}
Evaluating on the 150 training classes, our proposed models offer competitive results. Baseline flat metrics are used to compare the performance, as shown in Table \ref{table:eval_baseline}. Without surprise, the best performance is achieved by the \textit{Softmax} baseline, which agrees with the observation from \cite{frome2013devise}, classification formulations usually achieves higher accuracy than regression formulations. 
At the same time, our proposed models \textit{Joint-Cosine} and \textit{Word2Vec+} fall short of \textit{Softmax} by only around $1\%$, which is an affordable sacrifice given the zero-shot prediction capability and interpretability that will be discussed later. Visual results of the best proposed model \textit{Joint-Cosine} are shown in Figure \ref{fig:result_150}.

\vspace{-5px}
\subsubsection{Zero-shot predictions}
We then move to the zero-shot prediction tasks to fully leverage the hierarchical prediction ability of our models. The models are evaluated on 500 less frequent object classes in the ADE20K dataset. Predictions can be in the 500 classes, or their hypernyms, which could be compared based on our open vocabulary metrics.

\textit{Softmax} and \textit{Conditional Softmax} models are not able to make inferences outside the training classes, so we take their predictions within the 150 classes for evaluation. 

\textit{Convex Combination} \cite{norouzi2013zero} is another baseline model for comparison: we take the probability output from \textit{Softmax} within the 150 classes, to form new embeddings in the word vector space, and then find the nearest neighbors in vector space. This approach does not require re-training, but still offers reasonable performance.

Most of our proposed models can retrieve the hypernym of the testing classes, except \textit{*-Cosine} as they throw away the norm information during scoring, which is important for hypernym predictions.


\begin{figure}
\includegraphics[width=0.52\textwidth,trim={3cm 0 0.5cm 0},clip]{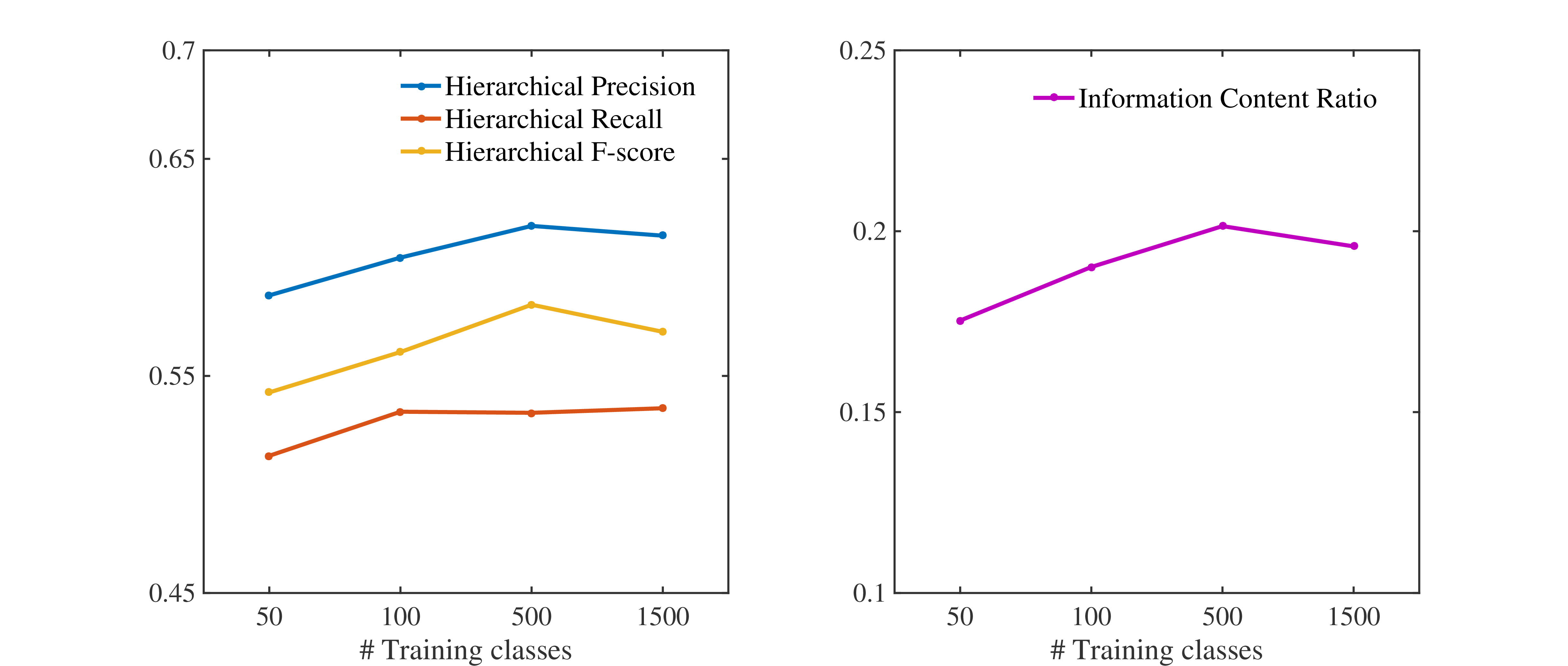}
\vspace{-5px}
\caption{Diversity test, evaluated with hierarchical metrics.}
\label{fig:diversity}
\vspace{-5px}
\end{figure}

Table \ref{table:eval_zeroshot} shows results on zero-shot predictions. In terms of the hierarchical metrics, \textit{Joint-Hyper} gives the best performance. And our proposed models in general win by a large margin over baseline methods. It confirms us that modeling the asymmetric relations of data pairs better represents the hierarchy. Figure \ref{fig:result_zeroshot} shows some prediction samples of our best model \textit{Joint-Hyper} (see Supplementary Materials for full predictions of our model). In each image, we only show one ground truth category to make things clear, different colors represent different predictions. Though the model does not always get the ground truth labels exactly correct, it gives reasonable predictions. Another observation is that predictions are sometimes noisy, we get 2-3 predictions on a single objects. Some of the inconsistencies are plausible though, \textit{e.g.} in the first row, the upper part of the ``rocking chair" is predicted as ``chair'' while the lower part is predicted as ``furniture''. As the pixels in the upper segment are \textit{closer} to ordinary chairs while the lower segment does not, so in the latter case the model gives a more general prediction.


\subsection{Diversity test}
\label{sec:exp2}
The open vocabulary recognition problem naturally raises a question: how many training classes do we need to generalize well on zero-shot tasks? To answer this question, we do a diversity test in this section. 

Different from the previous experiments, we do not take the most frequent classes for training, instead uniformly sample training and testing classes from the histogram of pixel numbers.
For better comparison, we fix the number of zero-shot test set classes to be 500, and the training classes range from 50 to 1500. In the training process, we offset the unbalance in pixel numbers by weighting the training class loss with their corresponding information content, so the less frequent classes contribute higher loss.

We only experiment with our best model \textit{Joint-Hyper} for this diversity test. Results in 
Figure \ref{fig:diversity} suggest that performance could saturate after training with more than 500 classes. We conjecture that training with many classes with few instances could introduce sample noise. So to further improve performance, more high quality data is required.
\begin{figure}
\centering
\includegraphics[width=0.5\textwidth]{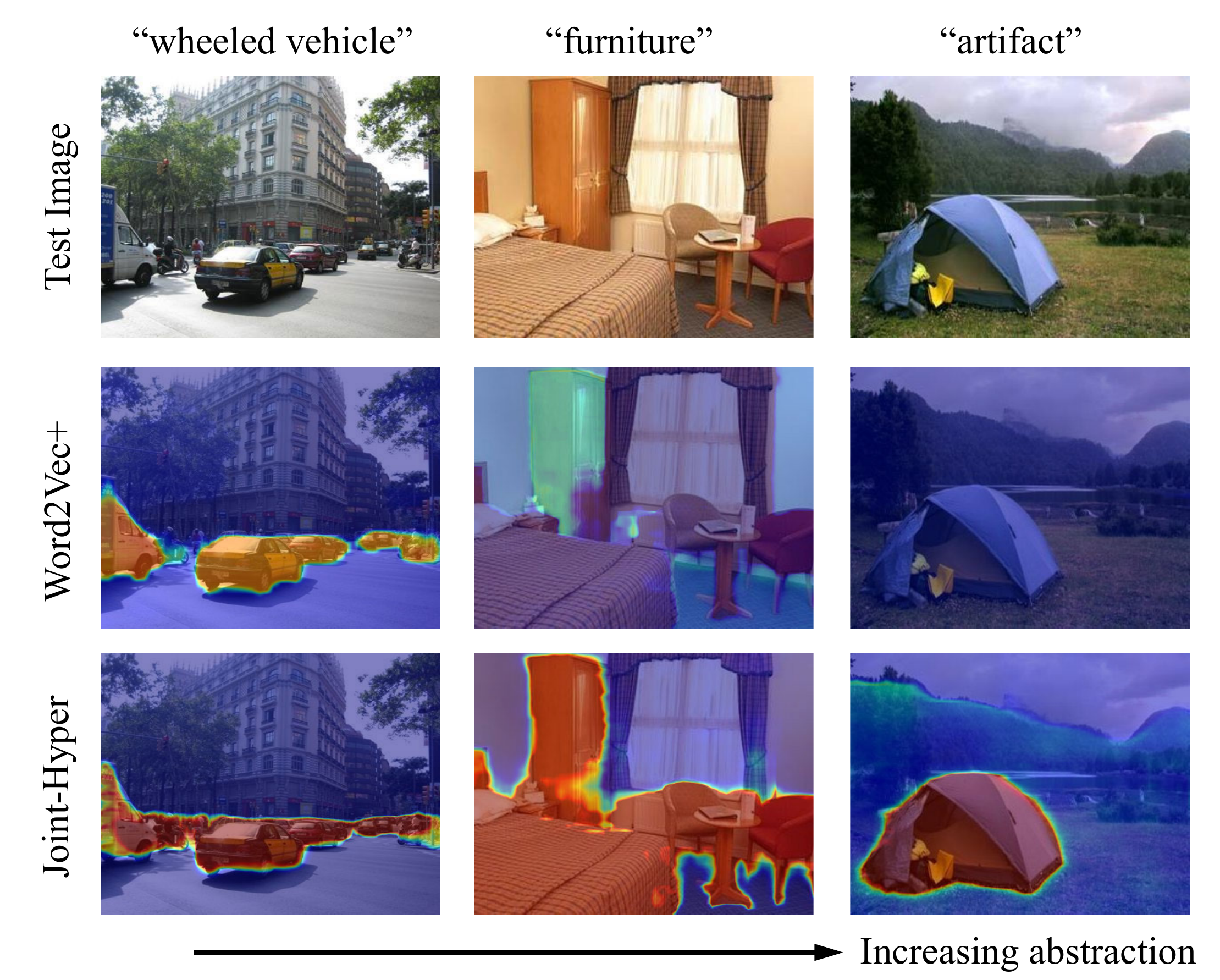}
\caption{Pixel-level concept search with increasing abstraction.}
\label{fig:retrieval}
\end{figure}

\begin{figure}
\centering
\includegraphics[width=0.48\textwidth]{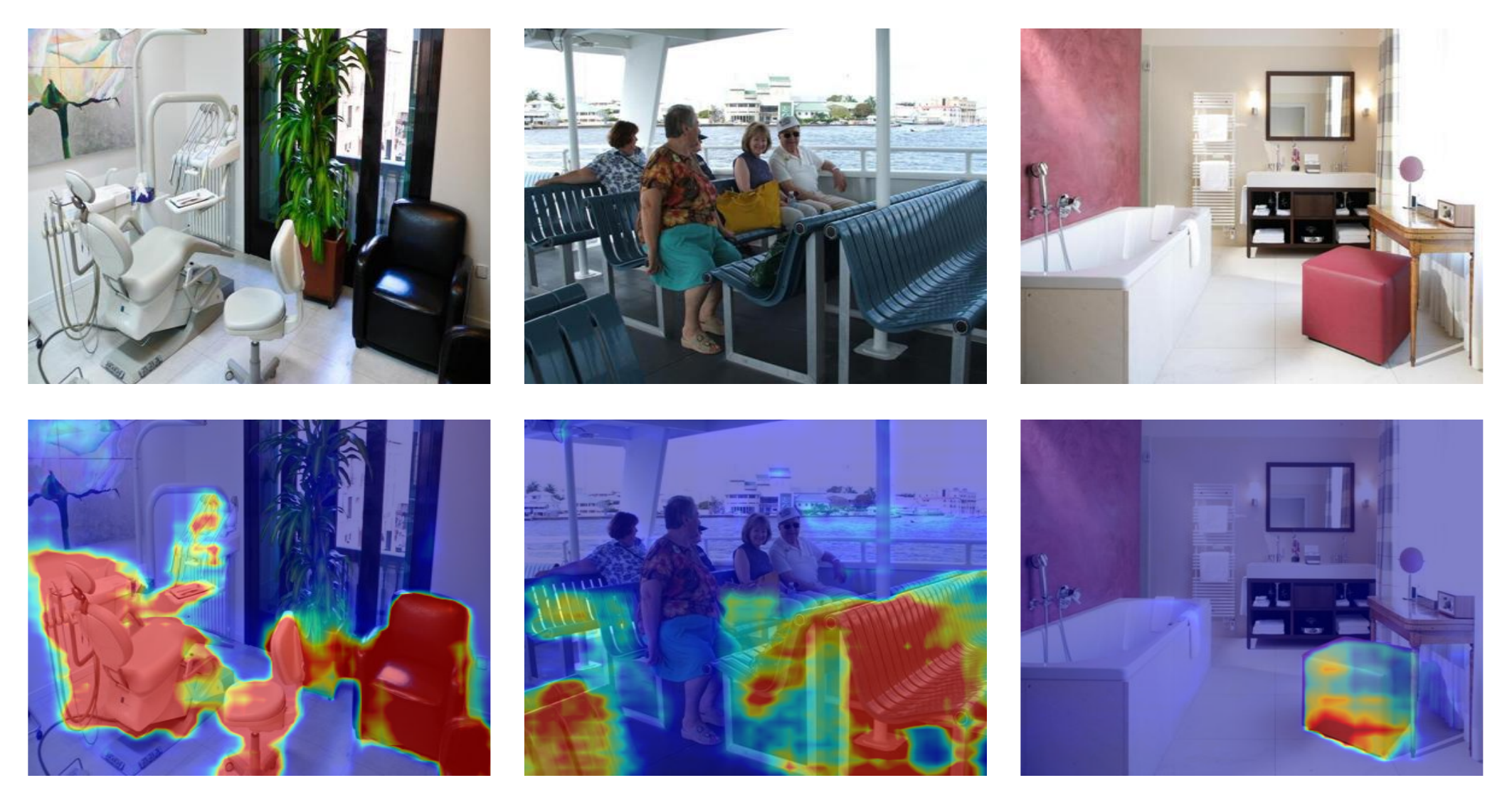}
\caption{Sittable objects have high scores while retrieving ``chair", indicating abstract attributes encoded in the embedding space.}
\label{fig:chairs}
\vspace{-5px}
\end{figure}

\begin{figure}
\centering
\includegraphics[width=0.5\textwidth]{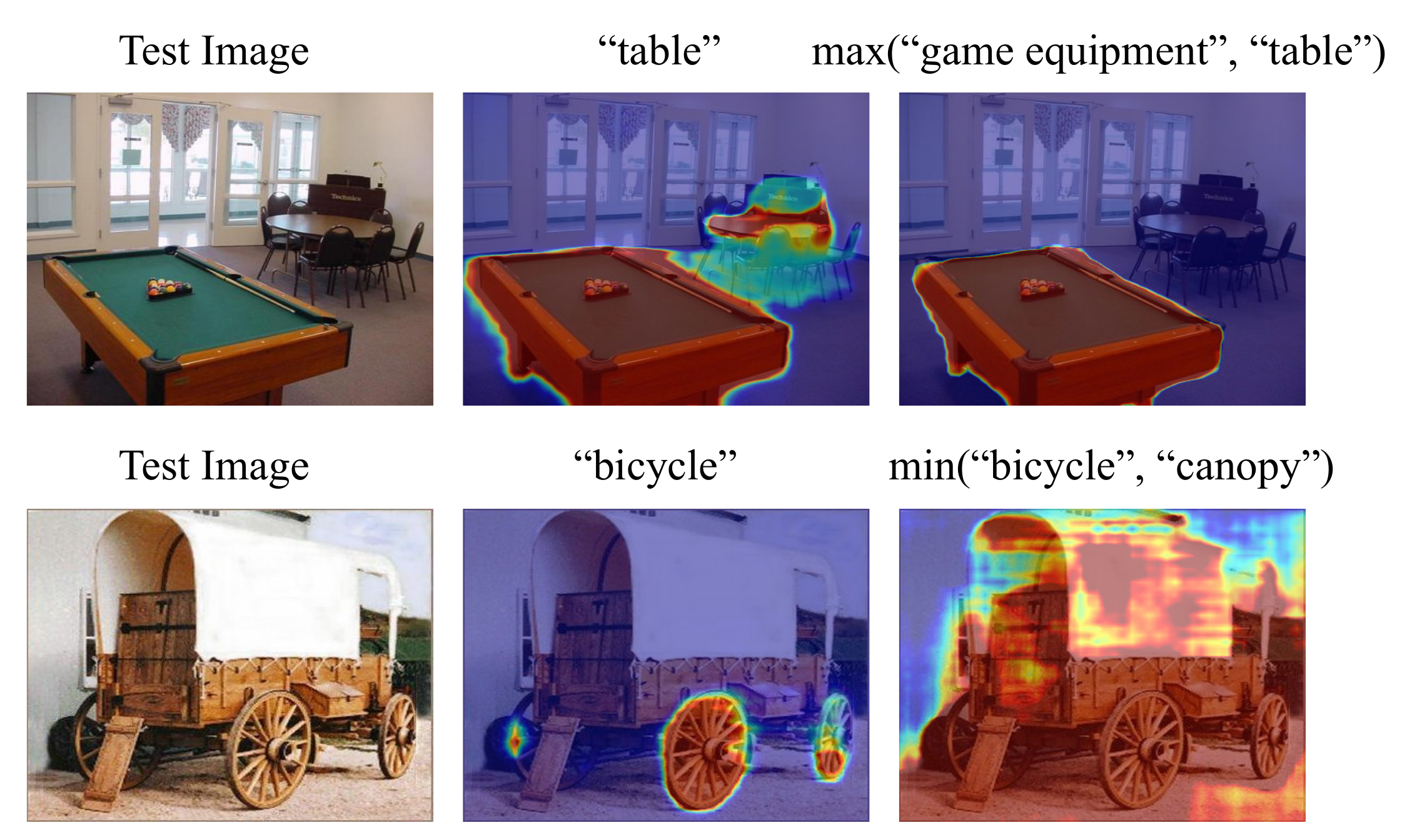}
\caption{Pixel-level search with synthesized concepts through arithmetic operations. Intersections and unions are achieved in the embedding space by \texttt{max} and \texttt{min}.}
\label{fig:arithmetics}
\vspace{-10px}
\end{figure}

\section{Interpreting the embedding space}
The joint embedding space we trained earlier features different properties from known spaces like \textit{Word2Vec}. In this section, we conduct three tests to explore them. 

\noindent\textbf{Concept search}.
In our framework, the joint training does not require all the concepts to have corresponding image data, the semantics can be propagated. This enables us to train with all the WordNet synsets and search with concepts that are not trained with images. In Figure \ref{fig:retrieval}, we show some pixel-level concept search results. The heatmaps are the scores in corresponding embedding spaces. As the search concepts become increasingly abstract, our model far outperforms \textit{Word2Vec+}, showing the effective encoding of hierarchical information in our embedding space.

\noindent\textbf{Implicit attributes encoding}.
One intriguing property of feature embeddings is that it is a continuous space, and classification boundaries are flexible. So we explore the vicinity of some concepts. In Figure \ref{fig:chairs}, we show score maps when searching for the concept ``chair". Interestingly, it is a common phenomenon that objects like ``bench'' and ``ottoman'', which are not hyponyms of ``chair" in WordNet, get reasonable response. We conjecture that the embedding space implicitly encodes some abstract attributes by clustering them, \textit{e.g.} \textit{sittable} is an affordance attribute. So by simply loosing classification threshold of ``chair'', one can detect regions where one can sit on.

\noindent\textbf{Synthesized concepts with arithmetics}.
Similar to \textit{Word2Vec}, in our joint embedding space, new concepts or image detectors can be synthesized with arithmetics. As shown in Figure \ref{fig:arithmetics}, we take elementwise \texttt{min} and \texttt{max} operations on the word concepts, and then search for these 
synthesized concepts in the images. It can be found that \texttt{max} operation takes the intersection of the concepts, \textit{e.g.} the pool table is the common hyponym of ``table" and ``game equipment"; and \texttt{min} takes the union, \textit{e.g.} the cart is composed of attributes of ``bicycle" and ``canopy". 

\section{Conclusion}
We introduced a new challenging task: open vocabulary scene parsing, which aims at parsing images in the wild. And we proposed a framework to solve it by embedding concepts from a knowledge graph and image pixel features into a joint vector space, where the semantic hierarchy is preserved. We showed our model performs well on open vocabulary parsing, and further explored the semantics learned in the embedding space.

\vspace{5px}
\noindent\textbf{Acknowledgement:} This work was supported by Samsung and NSF grant No.1524817 to AT. SF acknowledges the support from NSERC. BZ is supported by Facebook Fellowship. 

{\footnotesize
\bibliographystyle{ieee}
\bibliography{egbib}
}

\newpage
\section*{Supplementary Materials}

\section*{1. Data association protocol}
\begin{figure*}[h]
	\centering
	\includegraphics[width=1.05\linewidth] {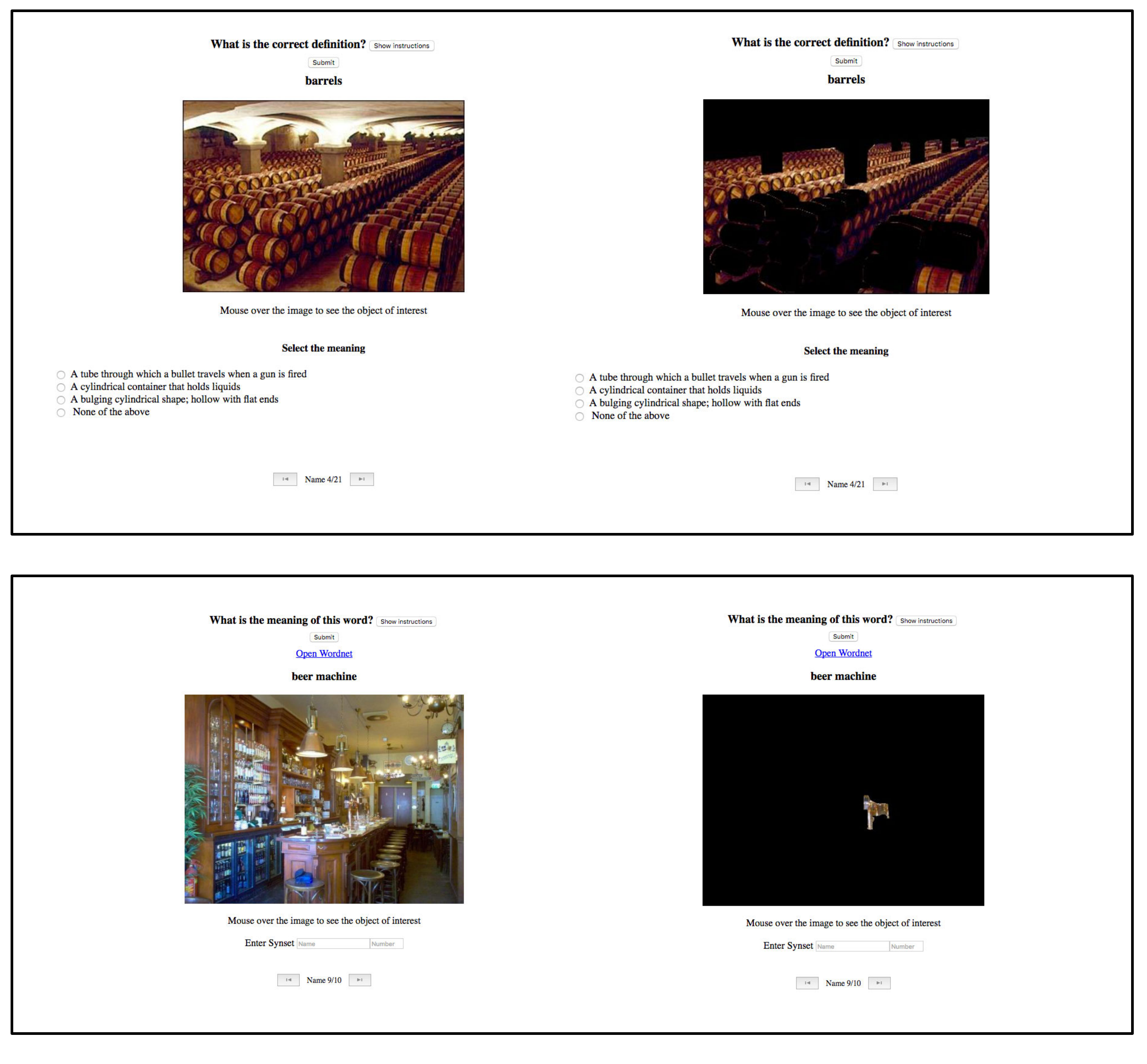}
	\vspace{2px}
	\caption{Screen shots of AMT interface for data association.}
	\label{fig:amt}
\end{figure*}

To learn the joint embeddings of images and word concepts, we need to augment ADE20K dataset by adding information about how the label classes ($>3000$) are semantically related. We associate each class in ADE20K dataset with a synset in WordNet, representing a unique concept. The data association process requires semantic understanding, so we resort to Amazon Mechanical Turk (AMT). The annotation protocol is detailed as follows, and screen shots of our AMT interface are shown in Figure \ref{fig:amt}. 

We search for each class in the dataset, for all the synsets having the same name. We find 3 different cases: (1) a single synset is found for the given class; (2) multiple synsets are found due to polysemy; (3) no sysnets are found, either because the correct synset has a different name or because that concept is not in WordNet. 

In the first case, we automatically match classes in the dataset with the obtained synsets, and then ask workers on AMT to verify by looking at the image labels and the definitions of synsets in the WordNet. 

In the second case where multiple synsets were found, we show an image displaying such concept and ask workers to select the synset whose definition matches the given class.

In the last case where no synset candidate was found, we show an image with the concept and ask workers to find the best matching synset by looking over WordNet online API. They also have the option to indicate when no synset can match.

\section*{2. Concept graph}
After data association, we end up getting 3019 classes in the dataset having synset matches. Out of these there are 2019 unique synsets forming a DAG. All the matched synsets have \textit{entity.n.01} as the top hypernym and there are in average 8.2 synsets in between. The depths of the ADE20K dataset annotations range from 4 to 19. 

\begin{figure*}[h]
	\centering
	\includegraphics[width=1.1\linewidth,trim={10cm 8cm 10cm 40cm},clip] {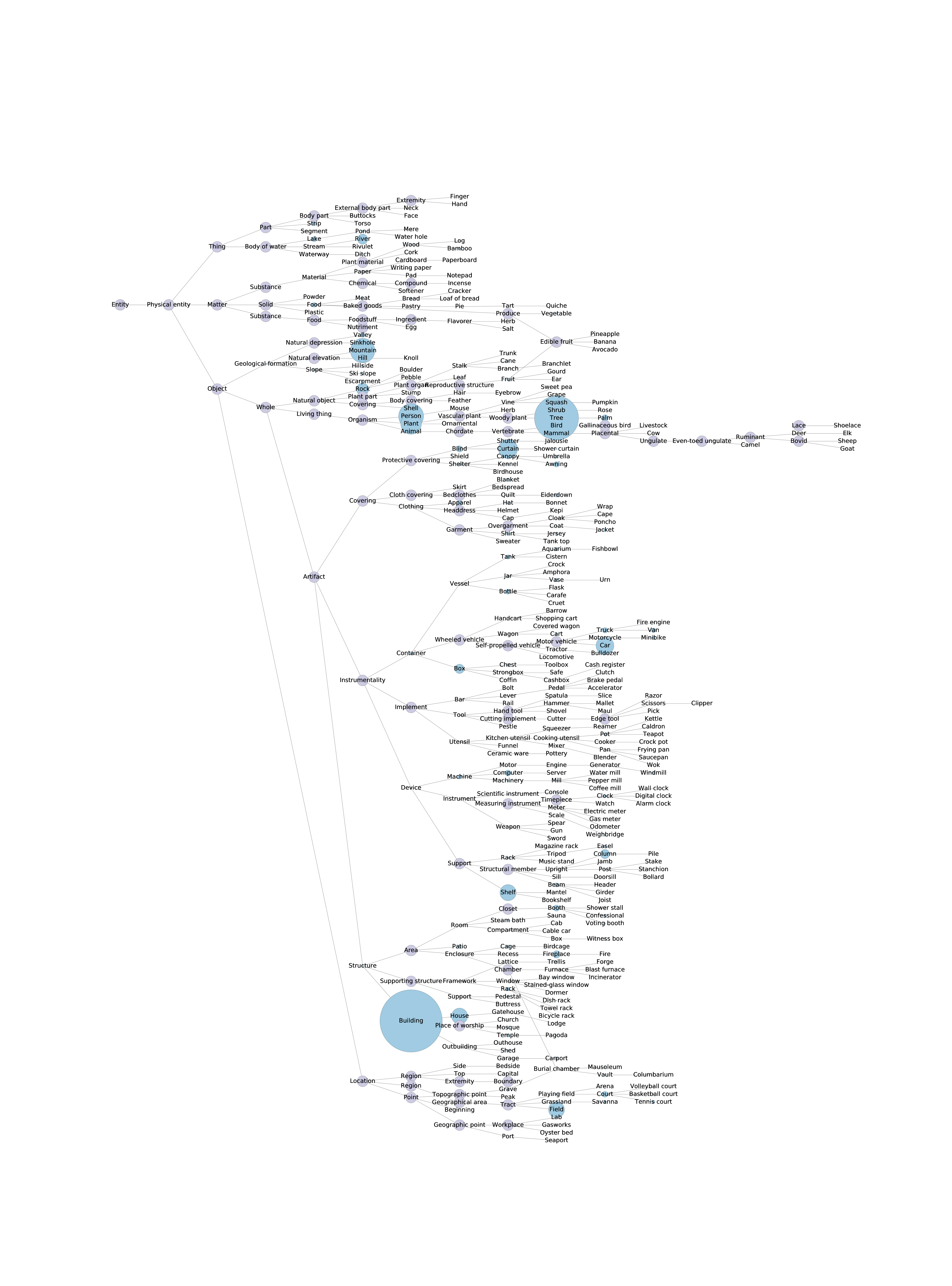}
	\vspace{-70px}
	\caption{Part of the concept graph built based on WordNet and ADE20K label frequencies.}
	\label{fig:dag}
\end{figure*}

A detailed visualization of the concept graph built is shown in Figure~\ref{fig:dag}. The node radii indicate the class frequencies in the ADE20K dataset. The figure only shows part of the full graph, nodes with 5 descendents or less have been hidden.

\vfill\null

\section*{3. Full zero-shot prediction lists}
\begin{figure*}[h]
	\centering
	\includegraphics[width=0.9\linewidth] {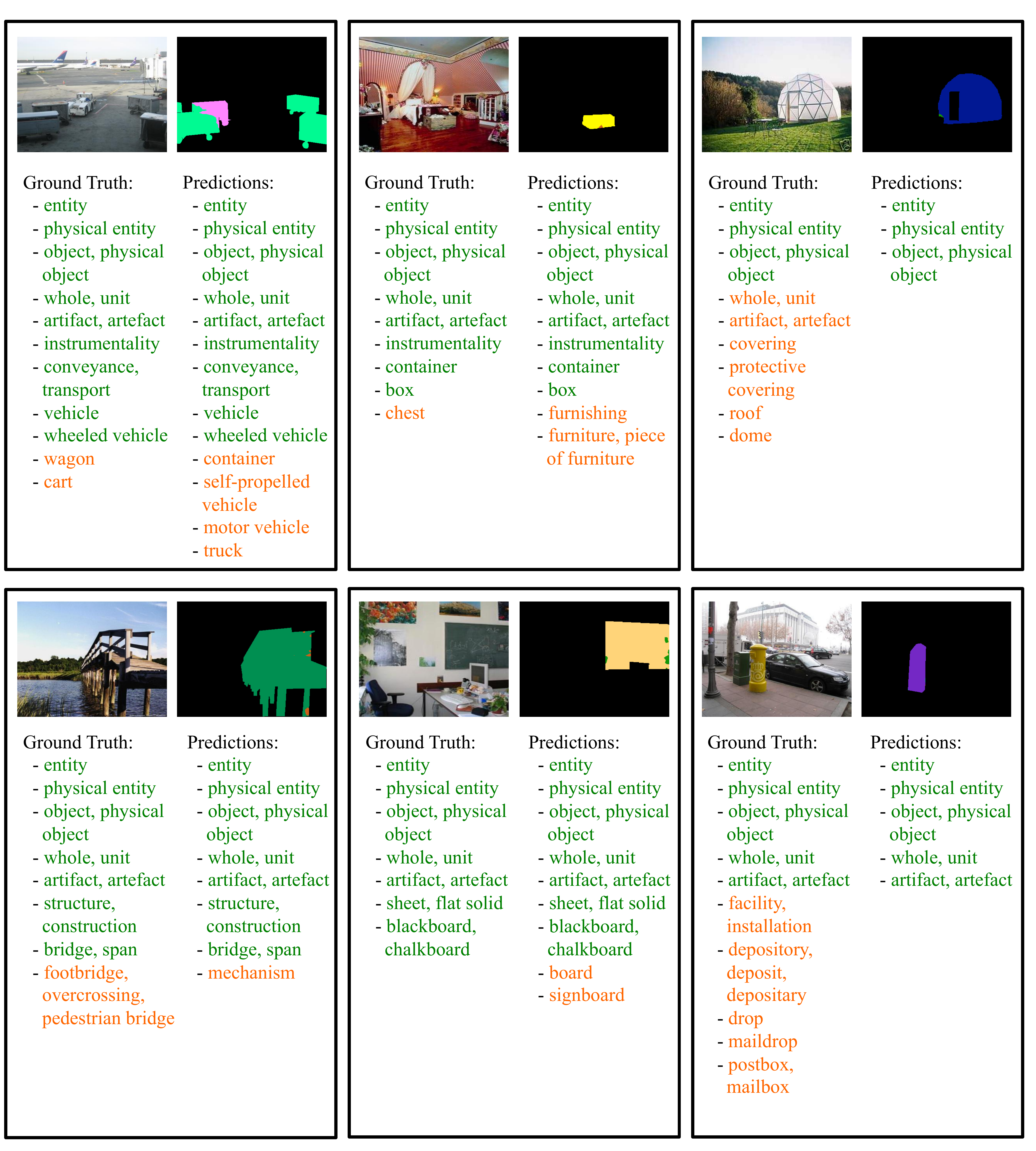}
	\vspace{4px}
	\caption{Full prediction results of zero-shot scene parsing.}
	\label{fig:zeroshot}
\end{figure*}

Our model gives each sample a list of predictions in hierarchical order. Due to the page limitations, full prediction lists are not shown in the main paper. In Figure~\ref{fig:zeroshot}, we give details of zero-shot predictions, both ground truth and prediction lists are shown in the texts beneath the images. Correct predictions are marked in green, inconsistent items are marked in orange. It can be seen that for hard examples, \textit{e.g.} ``dome" (row1, column3), a general and conservative prediction is made; when the test sample is easy and similar to training samples, \textit{e.g.} ``wagon" (row1, column1), our model gives specific and aggressive predictions.

\end{document}